\documentclass{article}
\usepackage[utf8]{inputenc}
\usepackage[T1]{fontenc}
\usepackage{geometry}
\usepackage{setspace}
\usepackage{titlesec}
\usepackage{natbib}
\usepackage{booktabs}
\usepackage{amsmath}
\usepackage{graphicx}
\usepackage{tabularx}
\usepackage{graphicx}

\title{Medical Reasoning in LLMs: An In-Depth Analysis of DeepSeek R1}
\author{Birger Moëll\,$^{1}$, Fredrik Sand Aronsson\,$^{2}$, Sanian Akbar\,$^{3}$  \\ 
$^{1}$KTH Royal Institute of Technology\\
$^{2}$Karolinska Institute\\
$^{3}$Stockholm Health Care Services, Region of Stockholm}
\date{\today}

\begin{document}

\maketitle

\begin{abstract}
The integration of large language models (LLMs) into healthcare holds immense promise, but also raises critical challenges, particularly regarding the interpretability and reliability of their reasoning processes. While models like DeepSeek R1—which incorporates explicit reasoning steps—show promise in enhancing performance and explainability, their alignment with domain-specific expert reasoning remains understudied. This paper evaluates the medical reasoning capabilities of DeepSeek R1, comparing its outputs to the reasoning patterns of medical domain experts. Through qualitative and quantitative analyses of 100 diverse clinical cases from the MedQA dataset, we demonstrate that DeepSeek R1 achieves 93\% diagnostic accuracy and shows patterns of medical reasoning. Analysis of the seven error cases revealed several recurring errors: anchoring bias, difficulty integrating conflicting data, limited consideration of alternative diagnoses, overthinking, incomplete knowledge, and prioritizing definitive treatment over crucial intermediate steps. These findings highlight areas for improvement in LLM reasoning for medical applications. Notably the length of reasoning was important with longer responses having a higher probability for error. The marked disparity in reasoning length suggests that extended explanations may signal uncer-
tainty or reflect attempts to rationalize incorrect conclusions. Shorter responses (e.g., under 5,000
characters) were strongly associated with accuracy, providing a practical threshold for assessing
confidence in model-generated answers. Beyond observed reasoning errors, the LLM demonstrated sound clinical judgment by systematically evaluating patient information, forming a differential diagnosis, and selecting appropriate treatment based on established guidelines, drug efficacy, resistance patterns, and patient-specific factors. This ability to integrate complex information and apply clinical knowledge highlights the potential of LLMs for supporting medical decision-making through artificial medical reasoning.
\end{abstract}

\section{Introduction}
\label{sec:introduction}
The accelerating adoption of artificial intelligence (AI) in healthcare, particularly large language models (LLMs), presents unprecedented opportunities to augment clinical decision-making and potentially improve patient outcomes. Clinical reasoning, the cornerstone of medical practice, is a complex cognitive process where practitioners integrate heterogeneous data streams, apply specialized knowledge frameworks, and navigate uncertainty to arrive at diagnostic and therapeutic decisions \citep{Jay2024, Sudacka2023}. This high-stakes process remains vulnerable to systemic failures, as evidenced by research suggesting medical errors contribute to over 250,000 deaths annually in the US, making it the third leading cause of death. Medical error includes unintended acts, execution failures, planning errors, or deviations from care processes that may cause harm \textbf{Makaryi2139}.

These challenges are exacerbated as healthcare systems worldwide face mounting pressures from workforce shortages \citep{who2023workforce} and increasing diagnostic complexity. In this strained environment, LLMs have emerged as potential aids to support clinical decision-making by potentially reducing cognitive burdens and mitigating error risks. However, the integration of these systems into medical workflows demands rigorous examination of their reasoning capabilities - not just their factual knowledge, but their ability to emulate the nuanced cognitive processes of expert clinicians while addressing systemic vulnerabilities in care delivery.

\subsection{Clinical Reasoning in Healthcare}
Clinical reasoning is an essential skill for healthcare professionals, particularly physicians \citep{DiazCrescitelli2019, durning2024teaching}. It encompasses all aspects of clinical practice, including patient management, treatment decisions, and ongoing care \citep{DiazCrescitelli2019}. While extensive research has focused on this area, challenges remain in understanding and implementing effective clinical reasoning \citep{yazdani2019five}.

A tension exists between explicit, quantitative approaches and the inherent limitations of human cognition, leading to the recognition that clinical reasoning involves both analytical and non-analytical processes, as described in dual-process theory \citep{pelaccia2011analysis, ferreira2010clinical}. Understanding how clinicians utilize both System 1 (intuitive) and System 2 (analytical) reasoning is crucial for evaluating whether LLMs can replicate this nuanced cognitive process.

\subsubsection{Theoretical Models and Cognitive Processes}
Several theoretical frameworks have shaped our understanding of clinical reasoning:

\begin{itemize}
    \item \textbf{Hypothetico-Deductive Reasoning:} Clinicians generate and test hypotheses using clinical data \citep{nierenberg2020using}. This model, while foundational, has been refined as research indicates clinical reasoning is more domain-specific and knowledge-dependent than initially thought.
    
    \item \textbf{Script Theory:} Medical knowledge is organized into "illness scripts"—cognitive frameworks that integrate clinical findings, risk factors, and pathophysiology \citep{Gee2017, charlin2000scripts}. Evaluating LLMs requires assessing their ability to form and utilize analogous script-like structures.
    
    \item \textbf{Dual Process Theory:} This influential framework describes two systems of thinking: a fast, intuitive system (Type 1) and a slower, analytical system (Type 2) \citep{gold2022clinical, custers2013medical}. Clinicians flexibly switch between these modes based on experience and situation \citep{boushehri2015clinical}. This highlights the need to evaluate LLMs on both rapid, pattern-recognition tasks and more complex, analytical scenarios.
    
    \item \textbf{Situated and Distributed Cognition:} Clinical reasoning is influenced by environmental factors, patient interactions, and team dynamics \citep{gold2022clinical, durning2011situativity}. Factors like fatigue and time pressure can impact the process \citep{torre2020widening}. This suggests that evaluating LLMs should consider their performance under various contextual constraints.
\end{itemize}

Clinical reasoning operates through both rapid, intuitive (System 1) and slower, analytical (System 2) cognitive processes. System 1 relies on pattern recognition and experience to generate immediate diagnostic hypotheses, while System 2 involves deliberate, systematic evaluation of information \citep{shimozono2020cognitive, chaves2022use}. Clinicians flexibly switch between these modes depending on case complexity \citep{Shimizu2012, Olupeliyawa2017}.

\subsubsection{Development of Expertise}
The development of clinical reasoning expertise involves a progression from deductive reasoning to the refinement of illness scripts, enabling more efficient diagnostic processes \citep{Shin2019, Radovic2022, lubarsky2015using}. This involves mastering data gathering, hypothesis generation, differential diagnosis, and management planning \citep{Weinstein2017}.  Assessing an LLM's ability to simulate this developmental trajectory could provide insights into its potential for clinical reasoning. 

\subsubsection{Diagnostic errors}
Diagnostic errors, often linked to reasoning failures, contribute significantly to preventable adverse events \citep{Mettarikanon2024, Zwaan2010}. Cognitive errors, particularly biases in information processing, are implicated in a majority of diagnostic errors \citep{graber2005diagnostic, Mukhopadhyay2024, schiff2013primary}. Common biases include representative heuristic, availability heuristic, and anchoring \citep{Kim2018}.  This underscores the importance of evaluating LLMs for susceptibility to similar cognitive biases.

Structured reflection and deliberate analysis can improve diagnostic accuracy \citep{moroz2017clinical}. However, the optimal balance between intuitive and analytical reasoning depends on various factors \citep{welch2017grounded}. This suggests that evaluating LLMs should involve tasks that require both rapid, intuitive responses and more deliberate, analytical reasoning.

The theoretical frameworks of clinical reasoning will inform the evaluation of DeepSeek R1 by providing a basis for analyzing its reasoning chains, identifying potential cognitive biases, and assessing its ability to navigate complex clinical scenarios analogous to human experts.

\subsection{Clinical Reasoning by LLMs}
The rapid evolution of LLMs presents both unprecedented opportunities and profound challenges for healthcare applications. While models like GPT-4 demonstrate remarkable performance on medical licensing examinations, achieving 87.6\% accuracy on USMLE-style questions \citep{nori2023capabilities}, performance metrics alone provide insufficient evidence for clinical deployment. Modern medicine requires reasoning that extends beyond factual recall to encompass contextual adaptation, probabilistic weighting of competing hypotheses, and adherence to evolving clinical guidelines \citep{rajpurkar2022ai}. A critical gap persists between LLMs' capacity to generate clinically plausible text and their ability to replicate the disciplined reasoning processes that underlie safe patient care \citep{singhal2023large}.  Reasoning models such as DeepSeek R1 \citep{deepseekai2025deepseekr1incentivizingreasoningcapability} output reasoning tokens, a chain of thought process of thinking in text before giving a text response. By evaluating reasoning tokens we can evaluate whether DeepSeek R1's \citep{deepseekai2025deepseekr1incentivizingreasoningcapability} reasoning aligns with that of medical experts, particularly in complex clinical scenarios.  DeepSeek R1 is designed to generate explicit inference chains through chain-of-thought prompting \citep{deepseekai2025deepseekr1incentivizingreasoningcapability}, offering a degree of interpretability that is crucial for medical applications. This paper focuses on DeepSeek R1 because its architecture, which emphasizes explicit reasoning steps, provides a unique opportunity to analyze the fidelity of its medical reasoning in comparison to human experts. The model is available open source which makes it possible to deploy on site for potential handling of sensitive clinical data.

The urgency of this research stems from the accelerating real-world deployment of medical LLMs despite unresolved limitations. A 2023 survey found 38\% of U.S. health systems piloting LLM-based tools \citep{HIMSS2024}, while regulatory approvals for AI diagnostics increased 127\% annually since 2020 \citep{Benjamens2020}. The potential risks of deploying LLMs without a thorough understanding of their reasoning abilities underscore the need for this research. Our work bridges critical gaps by:

\begin{itemize}
\item \textbf{Establishing validity metrics beyond answer correctness, focusing on medical reasoning ability.} We evaluate not just *what* the LLM answers, but *how* it arrives at that answer, analyzing the steps in its reasoning process. This goes beyond simple accuracy metrics to assess the quality and appropriateness of the reasoning itself.
\item \textbf{Identifying high-risk error patterns requiring mitigation, such as anchoring bias, protocol violations, and misinterpretations of lab values.}  Our analysis of DeepSeek R1's errors reveals specific cognitive biases and knowledge gaps that could lead to patient harm.  Identifying these patterns is crucial for developing mitigation strategies.
\item \textbf{Providing a foundation for medically-grounded architectures and training paradigms.} By understanding the strengths and weaknesses of current LLM reasoning, we can inform the design of future models that better align with clinical reasoning processes. This includes exploring techniques like retrieval augmented generation (RAG) and fine-tuning on medical reasoning data.
\end{itemize}

As LLMs transition from experimental tools to clinical assets, it is imperative for reasoning transparency equivalent to human practitioners. Through systematic evaluation of reasoning chain fidelity, we lay the groundwork for AI systems that complement rather than conflict with clinical judgment, harnessing LLMs' potential while safeguarding evidence-based medicine.

One key benefit of reasoning models over previous LLMs is the reasoning as a solution to the black box problem of LLM outputs \cite{wang2024augmentingblackboxllmsmedical}. By following the models reasoning we can evaluate their solutions and see what errors in thinking or knowledge led to incorrect outcomes. This has great potential both from a medical and a technical perspective. From a medical perspective, the information can be valuable if common LLM reasoning errors mimic errors that humans make. If so we can use LLM reasoning errors to understand how we can better train physicians to have robust medical reasoning skills. From a technical perspective, medical reasoning outputs and medical reasoning  errors can be used for reasoning fine-tuning and reinforcement learning training \cite{deepseekai2025deepseekr1incentivizingreasoningcapability} as well as understanding what data sources might need to be added to the model to improve performance.

By evaluation reasoning we get a more granular understanding of both what the model knows and doesn't know and its reasoning process and the errors within that reasoning process.

\section{Methodology}
\label{sec:methods}

\subsection{Dataset}

\subsubsection{Evaluation Corpus}
The study utilized 100 clinically diverse questions from the MedQA benchmark \citep{jin2021disease}, a rigorously validated dataset derived from professional medical board examinations across multiple countries. MedQA's questions follow the United States Medical Licensing Examination (USMLE) format, testing diagnostic reasoning through clinical vignettes requiring:

\begin{itemize}
\item Interpretation of patient histories and physical findings
\item Selection of appropriate diagnostic tests
\item Application of therapeutic guidelines
\item Integration of pathophysiology knowledge
\end{itemize}

\begin{table}[h]
\centering
\footnotesize
\begin{tabular}{|l|c|c|}
\hline
\textbf{Specialty} & \textbf{Number of Questions} & \textbf{Percentage} \\ \hline
Gynecology (OB\/GYN) & 6 & 6\% \\ \hline
Pediatrics & 7 & 7\% \\ \hline
Genetics & 7 & 7\% \\ \hline
Cardiology & 7 & 7\% \\ \hline
Neurology & 12 & 12\% \\ \hline
Hematology & 7 & 7\% \\ \hline
Gastroenterology & 7 & 7\% \\ \hline
Pulmonology & 4 & 4\% \\ \hline
Nephrology & 6 & 6\% \\ \hline
Urology & 3 & 3\% \\ \hline
Infectious Disease & 10 & 10\% \\ \hline
Oncology & 7 & 7\% \\ \hline
Surgery & 5 & 5\% \\ \hline
Dermatology & 3 & 3\% \\ \hline
Endocrinology & 5 & 5\% \\ \hline
Psychiatry & 3 & 3\% \\ \hline
Orthopedics & 2 & 2\% \\ \hline
Emergency Medicine & 3 & 3\% \\ \hline
Medical Ethics & 1 & 1\% \\ \hline
Biostatistics/Epidemiology & 3 & 3\% \\ \hline
Pharmacology & 2 & 2\% \\ \hline
ENT (Otolaryngology) & 4 & 4\% \\ \hline
Pathology & 2 & 2\% \\ \hline
Immunology & 1 & 1\% \\ \hline
Toxicology & 1 & 1\% \\ \hline
Metabolic Disorders & 2 & 2\% \\ \hline
Research Methods & 1 & 1\% \\ \hline
Physiology & 1 & 1\% \\ \hline
Patient Safety & 1 & 1\% \\ \hline
Neonatology & 1 & 1\% \\ \hline
\textbf{Total} & \textbf{100} & \textbf{100\%} \\ \hline
\end{tabular}
\caption{Distribution of Medical Questions by Specialty}
\label{tab:specialty_distribution}
\end{table}

Questions were selected through random sampling to ensure a cover of a range of specialties within medicine. The amount of questions (n=100) was selected to facilitate human analysis of reasoning outputs.

\subsection{Model Implementation}
We evaluated DeepSeek-R1 \citep{deepseekai2025deepseekr1incentivizingreasoningcapability}, a 671B parameter mixture of expert reasoning-enhanced language model built through a novel multi-stage training pipeline that combines reinforcement learning and fine-tuning on reasoning data. We used the DeepSeek-Reasoner model available through the DeepSeek API with default params.

\subsubsection{System Prompt}
 Please analyze this medical question carefully. Consider the relevant medical knowledge, clinical guidelines, and logical reasoning needed. Then select the single most appropriate answer choice. Provide your answer as just the letter (A, B, C, or D).

\subsection{Error Classification Protocol}
\begin{itemize}
\item \textbf{Step 1}: Ground truth alignment check
\begin{itemize}
\item Compare final answer to MedQA reference
\end{itemize}

\item \textbf{Step 2}: Reasoning chain decomposition
\begin{itemize}
\item Break down into diagnostic/treatment decision points
\item Map to clinical reasoning taxonomy
\end{itemize}

\item \textbf{Step 3}: Expert validation
\begin{itemize}
\item Clinician review all errors and compared them to medical reasoning best practice.
\end{itemize}
\end{itemize}

\section{Results}
\label{sec:errors}
Author S.A who is a active medical professional performed analysis of the medical reasoning of the model. Additional analysis focused on model performance and cognitive errors was done by authors B.M and F.S.
The model achieved an overall accuracy of 93\% on the 100 MedQA questions. Our analysis focused on the 7 cases where the model made an error to identify patterns and mechanisms of reasoning failures.

\subsection{Reasoning analysis by medical professional}

\subsubsection{Error Case 1: Neonatal Bilious Vomiting}
The model's reasoning is hampered by anchoring bias, difficulty integrating conflicting data, limited consideration of alternative diagnoses, overthinking, and a somewhat incomplete understanding of the embryology involved. It struggles to efficiently process the information and prioritize the most relevant clues, hindering its ability to confidently reach the correct diagnosis.
\subsubsection{Error Case 2: Respiratory Failure}
The model correctly identifies key information such as age, risk factors, recent surgery and findings in the pulmonary artery. It excessively focuses on histological composition and fibrous remodelling, leading it to weighing other options as more likely.
\subsubsection{Error Case 3: Acute Limb Ischemia}
Limb ischemia is correctly identified. The model recognizes atrial fibrillation as a key risk factor for arterial emboli, and discusses Rutherford classifications and possible interventions (surgery vs. thrombolysis). It emphasizes the urgency of revascularization and reasons that surgical thrombectomy should be done because the patient’s presentation suggests an embolic source and immediate threat to the limb. It incorrectly weighs the definitive treatment as the answer and skips the important "next" step of heparin drip.
\subsubsection{Error Case 4: Porphyria Cutanea Tarda (PCT)}
Recognizes porphyria cutanea tarda (PCT) based on photosensitive blistering, dark urine, and hyperpigmentation. It explains that treatment typically involves phlebotomy or low-dose hydroxychloroquine. It dismisses invasive or less relevant options (liver transplantation, thalidomide) and incorrectly concludes that hydroxychloroquine (alternative first line treatment) is the best next step, largely because the patient’s ferritin level is normal. Normally, a professional would reason that phlebomoty (first-line treatment) can induce remission even with normal iron stores and hydroxychloroquine is used if patient cannot tolerate phlebotomy. 
\subsubsection{Error Case 5: Enzyme Kinetics}Recognizes hexokinase and glucokinase properties as candidates for an enzyme found in most tissues that phosphorylates glucose. It also correctly identifies it as hexokinase rather than glucokinase, noting that hexokinase has a low Km (high affinity). However, it concludes that this enzyme also has a high Vmax, leading it to pick the incorrect answer (“Low X and high Y”). The LLM’s final reasoning step confuses hexokinase’s lower capacity (lower Vmax) with a higher capacity, thereby arriving at the wrong choice.
\subsubsection{Error Case 6: Preterm PDA Management}It rightly identifies the continuous murmur as PDA-related and distinguishes between drugs that keep the ductus open (prostaglandin E1) and those that close it (indomethacin). However, it overestimates how age limits indomethacin’s use, leading it prematurely to favor surgical ligation. In actual clinical practice, a stable 5-week-old would still warrant a trial of pharmacologic closure before considering surgical options.

\subsubsection{Error Case 7: Niacin Flushing}
Correctly identifies that the patient experiences niacin-induced flushing after statin intolerance. It recognizes niacin as a likely cause of her evening flushing and pruritus, and appropriately considers—but rules out—alternative explanations such as carcinoid syndrome and pheochromocytoma, given hints of cancer in the patient's history. However, it departs from a typical medical approach by concluding that switching to fenofibrate (which primarily targets elevated triglycerides rather than LDL) is the best next step, rather than attempting to mitigate the flushing (for example, with NSAIDs) while maintaining niacin therapy. This oversight highlights a gap in its reasoning compared to standard clinical practice, where controlling niacin’s side effects is usually preferred before abandoning a therapy that addresses the patient’s elevated LDL cholesterol.

\begin{table}[h!]
\scriptsize
\centering
\caption{Summary of Reasoning Errors Across Cases}
\label{tab:error_summary}
\begin{tabular}{lllp{4cm}}
\toprule
\textbf{Case} & \textbf{Error Type} & \textbf{Model Answer} & \textbf{Key Reasoning Flaw} \\
\midrule
E1. Neonatal Vomiting & Anchoring Bias & B (Duodenal Atresia) & Overprioritized textbook presentation despite incompatible timeline \\
E2. Respiratory Failure & Etiology Confusion & C (Pulmonary Hypertension) & Misattributed vascular remodeling to primary disease \\
E3. Limb Ischemia & Protocol Violation & C (Surgery) & Skipped anticoagulation step in Rutherford IIb \\
E4. PCT Management & Lab Misinterpretation & D (Hydroxychloroquine) & Overvalued serum ferritin over hepatic iron \\
E5. Enzyme Kinetics & Isoform Confusion & C (High Vmax) & Confused hexokinase/glucokinase kinetic profiles \\
E6. PDA Management & Therapeutic Window Error & C (Surgery) & Misjudged indomethacin efficacy in preterms \\
E7. Niacin Flushing & Overinvestigation & D (Fenofibrate) & Ignored temporal drug-effect relationship \\
\bottomrule
\end{tabular}
\end{table}

\subsection{Detailed Error Analysis}

\subsubsection*{Error Case 1: Neonatal Bilious Vomiting}
\begin{itemize}
\item \textbf{Pathway of reasoning}:
\begin{equation*}
\scriptsize
\text{Bilious Vomit} \rightarrow \underbrace{\text{Duodenal Atresia}}_{\text{Model's Focus}} \rightarrow \text{Emergency Laparotomy} \leftarrow \text{Annular Pancreas} \leftarrow \text{Delayed Presentation + Normal Prenatal US}
\end{equation*}
\item \textbf{Critical Failure}: Anchoring bias on classic duodenal obstruction pattern while ignoring:
\begin{enumerate}
\item 3-week delayed presentation (incompatible with complete atresia)
\item Absence of prenatal ultrasound findings
\end{enumerate}
\item \textbf{Clinical Impact}: Risk of delayed annular pancreas diagnosis (24-48hr window for surgical intervention)
\end{itemize}

\subsubsection*{Error Case 2: Respiratory Failure}
\begin{itemize}
\item \textbf{Pathway of reasoning}:
\begin{equation*}
\text{DVT} \rightarrow \text{PE} \rightarrow 
\text{Fibrosis} \rightarrow  \text{Actual Cause} \rightarrow
\underbrace{\text{CTEPH}}_{\text{Model's Focus}}
\end{equation*}
\item \textbf{Critical Failure}: Attributed wall remodeling (effect) as primary pathology
\item \textbf{Risk Amplification}: Increased mortality from missed vasculitis diagnosis
\end{itemize}

\subsubsection*{Error Case 3: Acute Limb Ischemia}
\begin{itemize}
\item \textbf{Pathway of Reasoning}:
\begin{equation*}
\text{Ischemic Limb} \rightarrow \underbrace{\text{Direct Surgery}}_{\text{Model's Focus}} \rightarrow \text{Reperfusion Injury} \leftarrow \text{Heparin Bridge} \leftarrow \text{Imaging Guidance}
\end{equation*}
\item \textbf{Critical Failure}: Bypassed essential anticoagulation and imaging steps
\item \textbf{Risk Amplification}: Increased limb loss probability with delayed anticoagulation
\end{itemize}

\subsubsection*{Error Case 4: Porphyria Cutanea Tarda (PCT)}
\begin{itemize}
\item \textbf{Pathway of Reasoning}:
\begin{equation*}
\text{PCT} \rightarrow \text{Phlebotomy Required} \rightarrow \underbrace{\text{Normal Iron Stores }}_{\text{Model's Focus}}  \rightarrow \text{Hydroxychloroquinine} 
\end{equation*}
\item \textbf{Critical Failure}: Equated serum ferritin with total body iron stores
\item \textbf{Risk Amplification}: Increased risk of cirrhosis from persistent iron overload
\end{itemize}

\subsubsection*{Error Case 5: Enzyme Kinetics}
\begin{itemize}
\item \textbf{Pathway of Reasoning}:
\begin{equation*}
\scriptsize
\text{Tissue Distribution} \rightarrow \underbrace{\text{Low Vmax Assumption}}_{\text{Model's Focus}} \rightarrow \text{Metabolic Dysregulation} \leftarrow \text{Hexokinase Signature} \leftarrow \text{Low Km/High Vmax}
\end{equation*}
\item \textbf{Critical Failure}: Confused hexokinase (high-affinity/high-capacity) with glucokinase kinetics
\item \textbf{Risk Amplification}: Error in predicting glucose utilization rates
\end{itemize}

\subsubsection*{Error Case 6: Preterm PDA Management}
\begin{itemize}
\item \textbf{Pathway of Reasoning}:
\begin{equation*}
\text{Preterm Birth} \rightarrow \text{PDA} \rightarrow \underbrace{\text{Surgical Ligation}}_{\text{Model's Focus}} \leftarrow \text{Indomethacin Window} \leftarrow \text{5-Week Age}
\end{equation*}
\item \textbf{Critical Failure}: Overestimated surgical urgency in stable infant
\item \textbf{Risk Amplification}: Higher complication rate vs medical management
\end{itemize}

\subsubsection*{Error Case 7: Niacin Flushing}
\begin{itemize}
\item \textbf{Pathway of Reasoning}:
\begin{equation*}
\text{Niacin Use} \rightarrow \text{Flushing} \rightarrow \underbrace{\text{Fenofibrate Switch}}_{\text{Model's Focus}} \xleftarrow{\text{PGD2 Pathway}} \text{Aspirin Prophylaxis}
\end{equation*}
\item \textbf{Critical Failure}: Misattributed prostaglandin-mediated flushing to rare neoplasms
\item \textbf{Risk Amplification}: Reduced lipid control efficacy with unnecessary agent switch
\end{itemize}

\begin{table}[h!]
\centering
\caption{Distribution of Reasoning Errors in 100 Clinical Cases}
\label{tab:error_dist}
\begin{tabular}{lccc}
\toprule
\textbf{Error Type} & \textbf{Count} & \textbf{Percentage} & \textbf{Exemplar Case} \\
\midrule
Protocol Misapplication & 2 & 2\% & Acute limb ischemia management \\
Anchoring Bias & 1 & 1\% & Neonatal bilious vomiting \\
Etiology-Consequence Confusion & 1 & 1\% & Pulmonary artery fibrosis \\
Lab Value Overinterpretation & 1 & 1\% & Porphyria cutanea tarda \\
Isoform Misunderstanding & 1 & 1\% & Enzyme kinetics \\
Overinvestigation Tendency & 1 & 1\% & Niacin-induced flushing \\
\bottomrule
\end{tabular}
\end{table}

\subsection{Analysis of Diagnostic Reasoning Errors}
We found recurring patterns of diagnostic reasoning errors.  A key finding across multiple cases was \textbf{anchoring bias}, with fixation on an initial diagnosis (e.g., duodenal atresia in Case 1, CTEPH in Case 2) and subsequently failed to adequately incorporate conflicting evidence.  This was often compounded by \textbf{confirmation bias}, with selectively attending to information supporting the initial impression while dismissing contradictory data (e.g., normal ferritin in the context of suspected PCT in Case 4).

Several cases demonstrated errors related to disease pathway understanding. In Case 2, \textbf{feature binding} led to misattributing wall remodeling as the primary pathology rather than recognizing it as a consequence of another underlying condition (vasculitis).  A similar error in Case 5 involved confusing enzyme kinetics, misidentifying hexokinase as glucokinase, highlighting a lack of understanding of the specific biochemical pathways.

\textbf{Omission bias} was evident in Case 3, where crucial steps like anticoagulation and imaging were bypassed in the rush to surgery for acute limb ischemia.  This suggests a failure to consider all necessary elements of the diagnostic and treatment pathway.  In contrast, Case 6 demonstrated potential \textbf{commission bias} with the overestimation of surgical urgency in a stable infant with a PDA, potentially exposing the patient to unnecessary risk.

Finally, Case 7 illustrated an error in attribution, misattributing niacin-induced flushing to rare neoplasms instead of recognizing it as a prostaglandin-mediated effect.  This misattribution led to an unnecessary and detrimental change in lipid-lowering medication.

These findings emphasize the importance of recognizing and mitigating cognitive biases and ensuring a thorough understanding of disease pathways to improve diagnostic accuracy and patient safety.  The quantified risk amplifications associated with each error underscore the potential clinical impact of these reasoning flaws.

Another error we think is important to address is the one found in the first Case E1. If you follow the reasoning trace of the model it actually decides on A Abnormal migration of ventral pancreatic bud (correct) but outputs B, Complete failure of proximal duodenum to recanalize (false) . The model first reason and then outputs the answer. Although this only happened a single time, we want to highlight this because it shows that the reasoning might differ from the response. This means that in a clinical setting it is wise to have both model reasoning and model output in order to minimize the risk of errors. If a clinician would have access to both reasoning and output, the reasoning might help the clinician find the right diagnosis but having only access to the model output would lead to a potential misdiagnosis. This highlights the benefit or R1, which shows reasoning patterns, which are hidden in similar reasoning models such as O1 and O3 made by Open AI.

\begin{table}[h]
    \centering
    \scriptsize
    \renewcommand{\arraystretch}{1.2}
    \setlength{\tabcolsep}{4pt}
    \hspace*{-2cm} 
    \begin{tabular}{p{2.0cm} p{6cm} p{6cm} p{2.0cm} p{2.5cm}}
        \toprule
        \textbf{Question} & \textbf{Strengths} & \textbf{Weaknesses} & \textbf{Diagnosis} & \textbf{R1 Answer} \\
        \midrule
        \textbf{C1. 23-year-old pregnant woman with UTI} & 
        - Identifies cystitis based on symptoms. \newline
        - Recognizes need for treatment. \newline
        - Rules out inappropriate options. \newline
        - Selects Nitrofurantoin. &
        - Spends time on Cephalexin. \newline
        - Could be more concise. & 
        Cystitis & Cystitis \newline \textbf{Correct} \\
        \midrule
        \textbf{C2. 3-month-old with SIDS} & 
        - Correctly identifies SIDS. \newline
        - Recalls prevention strategies. \newline
        - Evaluates answer choices. \newline
        - Recognizes "Back to Sleep" campaign. & 
        - None significant. & 
        SIDS & SIDS \textbf{Correct} \\
        \midrule
        \textbf{C3. 20-year-old woman with menorrhagia} & 
        - Interprets lab results. \newline
        - Considers differentials. \newline
        - Recognizes family history. \newline
        - Identifies vWD. &
        - Briefly considers Hemophilia A. \newline
        - Mentions bleeding time. & 
        Von Willebrand disease & Von Willebrand disease \newline \textbf{Correct} \\
        \midrule
        \textbf{C4. 40-year-old zookeeper with pancreatitis} & 
        - Recalls causes of pancreatitis. \newline
        - Identifies scorpion sting. \newline
        - Considers other options. &
        - None significant. & 
        Scorpion sting & Scorpion sting \newline \textbf{Correct} \\
        \midrule
                \textbf{E1. 3-week-old with bilious vomiting} & 
        - Recognizes bilious vomiting as obstruction. \newline
        - Considers relevant differentials. \newline
        - Understands embryology. &
        - Initially rules out duodenal atresia. \newline
        - Fixates on "complete" in option B. \newline
        - Overemphasizes malrotation. \newline
        - Repetitive explanation. & 
        Abnormal migration of ventral pancreatic bud & 
        Duodenal atresia \newline \textbf{Incorrect} The models reason correctly but gives out the wrong response \\
        \midrule
           \textbf{E2. 58-year-old woman post-surgery} & 
        - Identifies risk factors. \newline
        - Initially leans towards thromboembolism. \newline
        - Considers each option. \newline
        - Understands CTEPH. &
        - Gets fixated on histological composition. \newline
        - Repetitive reasoning. & 
        Thromboembolism & Pulmonary Hypertension \newline \textbf{Incorrect} \\
        \midrule
         \textbf{E3. 68-year-old man with leg pain} & 
        - Correctly identifies acute limb ischemia. \newline
        - Recognizes atrial fibrillation as a risk factor. \newline
        - Applies Rutherford classifications to evaluate severity. \newline
        - Understands that urgent management is needed to salvage limb. & 
        - Incorrectly prioritizes definitive treatment over immediate anticoagulation with heparin. \newline
        - Incorrectly states that thrombolysis is contraindicated in embolic events. & 
         Heparin drip  & Surgical thrombectomy \newline
         \textbf{Incorrect}\\
        \midrule
         \textbf{E4. 48-year-old woman with photosensitive rash} & 
        - Correctly identifies porphyria cutanea tarda (PCT) as the most likely diagnosis. \newline
        - Recognizes the significance of family history, dark urine, and photosensitivity. \newline
        - Considers other porphyrias (variegate porphyria). \newline
        - Appropriately rules out liver transplantion and thalomide as standard therapies, understands the role of phlebotomy and hydroxychloroquine in PCT treatment. & 
        -Places excessive emphasis on normal ferritin levels, overlooking that phlebotomy can still induce remission even with normal iron stores. \newline
        - Briefly considers unrelated conditions (epidermolysis bullosa, pseudoporphyria). \newline
        - Incorrectly states that thalidomide is used in refractory cases of PCT. & 
        Begin phlebotomy therapy & Begin oral hydroxychloroquine therapy \newline \textbf{Incorrect} \\
        \midrule
        \textbf{E5. Enzyme Kinetics} &
        - Correctly relates X to Km and Y to Vmax. \newline
        - Correctly identifies the enzyme as hexokinase. \newline
        - Understands the properties of hexokinase (low Km). \newline
        - Correctly identifies that the enzyme in question phosphorylates glucose. &
        - Overthinks the Vmax, failing to definitively conclude whether it's high or low, causing confusion in the final step.\newline 
        -Confuses the concepts of Vmax and Km, incorrectly stating that a low Km indicates a high Vmax. \newline
        - Incorrectly states that hexokinase has a higher Vmax than glucokinase and incorrectly states that hexokinase is inhibited by glucose-6-phosphate under these experimental conditions.\newline
        - It overthinks minor details and loses track of the simpler hallmark difference &
        Low X and low Y & Low X and high Y \\ \textbf{Incorrect} & \\
        \midrule
        \textbf{E6. 5-week-old infant with a murmur} &
        - Correctly identifies PDA as the most likely diagnosis. \newline
        - Recognizes the significance of preterm birth. \newline
        - Understands the implications of the continuous murmur. \newline
        - Considers the infant's age and feeding changes. \newline
        - Knows the general management options for PDA (Indomethacin, surgery). &
        - Incorrectly dismisses indomethacin as an option based on age alone without considering the full clinical picture \newline
        - Overthinks the feeding changes and weight gain. \newline
        - Overthinks age and arrives at the wrong first-line treatment in an otherwise stable infant. &
        Indomethacin infusion & Surgical ligation \newline \textbf{Incorrect} \\
        \midrule
        \textbf{E7. 53-year-old woman with flushing and itching} &
        - Correctly identifies niacin-induced flushing as the most likely cause. \newline
        - Considers other possibilities (carcinoid, pheochromocytoma, allergy). \newline
        - Understands the limitations of statins and fibrates. \newline
        - Recognizes the need for LDL management. &
        - Incorrectly prioritizes switching to fenofibrate over managing niacin side effects. \newline
        - Overly focuses on the possibility of carcinoid syndrome despite the low likelihood. \newline
        - Fails to recognize that taking aspirin 30 minutes before niacin can significantly reduce flushing. &
        Administer ibuprofen & Switch niacin to fenofibrate \newline \textbf{Incorrect} \\
        \bottomrule

    \end{tabular}
    \caption{Examples of responses with a focus on incorrect responses and reasoning}
    \label{tab:evaluation}
\end{table}

\subsection{Statistical Analysis of Reasoning Lengths in Correct vs. Incorrect Responses}
\begin{figure}[h]
    \centering
    \includegraphics[width=1\textwidth]{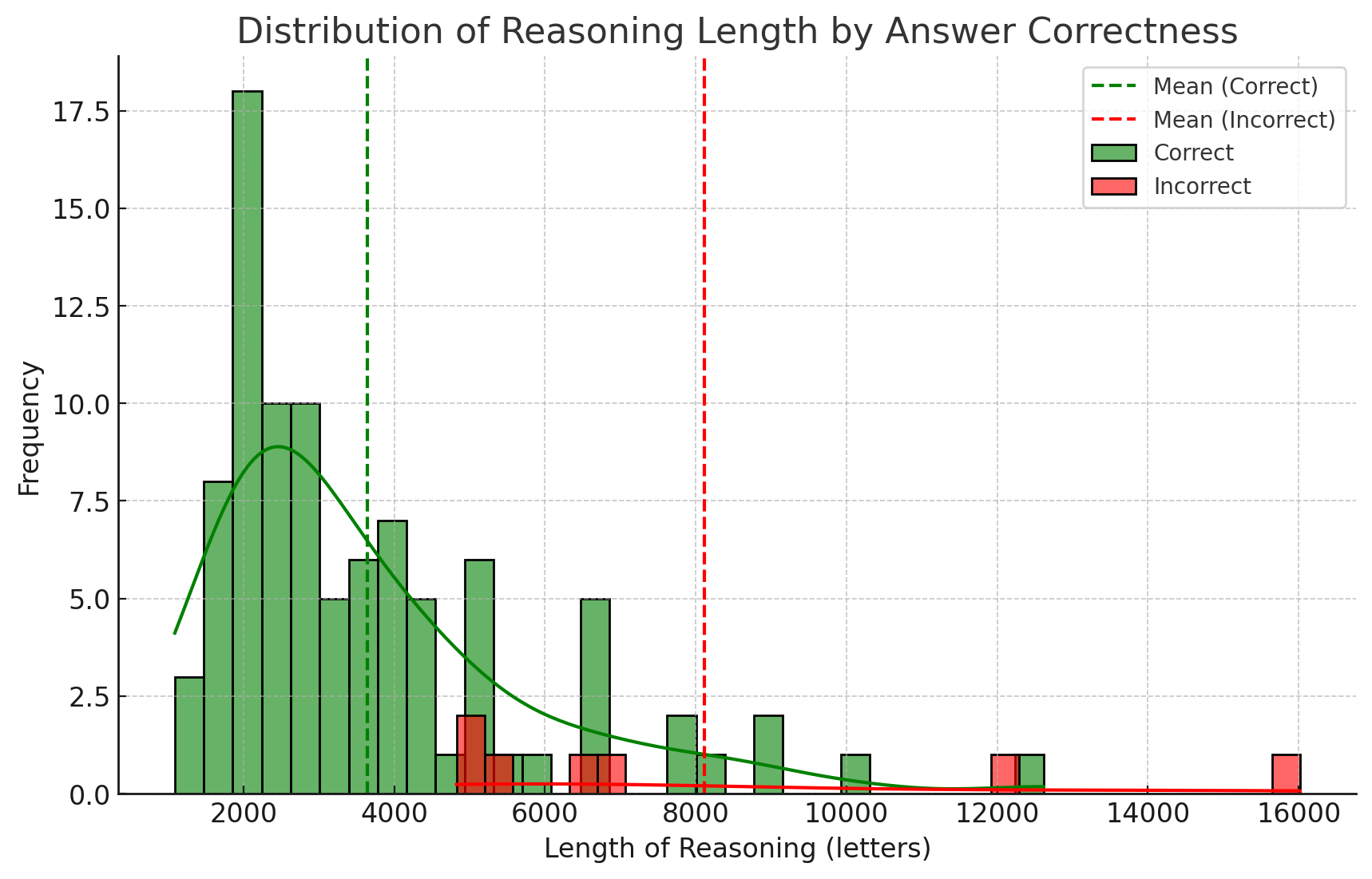}
    \caption{Length of reasoning and correctness.}
    \label{fig:example}
\end{figure}
We conducted an independent two-sample Welch’s t-test to compare the average reasoning length between correct and incorrect answers, as the groups exhibited unequal variances. The analysis revealed a statistically significant difference (t = -2.74, p = 0.032), with incorrect answers containing substantially longer reasoning (mean = 8,118 characters) compared to correct answers (mean = 3,648 characters). The negative t-value reflects the directional difference, where incorrect responses were consistently lengthier.

The marked disparity in reasoning length suggests that extended explanations may signal uncertainty or reflect attempts to rationalize incorrect conclusions. Shorter responses (e.g., under 5,000 characters) were strongly associated with accuracy, providing a practical threshold for assessing confidence in model-generated answers. This metric could enhance user transparency by flagging verbose outputs as potential indicators of unreliability.

\subsection{Analysis of reasoning success}
Although our effort focused on reasoning errors in most cases the model was successful with 93\% accuracy. In our analysis of the successful cases we found that the medical reasoning of the model was sound.

\subsubsection{Classification as Medical Reasoning}
The reasoning by the R1 model would likely qualify as medical reasoning. The thought process demonstrates key elements of clinical decision-making demonstrated here on case C1 (see table 4):

\subsubsection{Correct Case 1: A 23-year-old pregnant women at 22 weeks gestation presents with burning upon urination}
The model identifies that the patient is a pregnant woman at 22 weeks gestation with signs of a lower urinary tract infection. It systematically evaluates the safety and efficacy of each antibiotic option in pregnancy: it rules out ampicillin due to common resistance, ceftriaxone because it is overly broad for simple cystitis, and doxycycline because it is contraindicated in pregnancy. It concludes that nitrofurantoin is safe and effective in the second trimester, making option D the correct choice.

\begin{itemize}
    \item \textbf{Data synthesis:} Systematically reviews the patient’s history, symptoms, and exam findings.
    \item \textbf{Differential diagnosis:} Rules out pyelonephritis (absence of CVA tenderness) and narrows to cystitis.
    \item \textbf{Application of guidelines:} Considers pregnancy-specific risks and antibiotic safety profiles.
    \item \textbf{Critical appraisal of options:} Evaluates drug efficacy, resistance patterns, and contraindications.
    \item \textbf{Risk-benefit analysis:} Balances fetal safety (e.g., avoiding doxycycline) with maternal treatment efficacy.
\end{itemize}

\subsubsection{Structured Clinical Approach}
\begin{itemize}
    \item \textbf{Begins with clinical context:} Identifies pregnancy as a critical factor influencing management.
    \item \textbf{Prioritizes diagnosis:} Distinguishes cystitis from pyelonephritis based on exam findings (no CVA tenderness).
    \item \textbf{Antibiotic stewardship:} Avoids overly broad agents (ceftriaxone) for uncomplicated cystitis and considers resistance patterns (ampicillin’s limitations).
    \item \textbf{Guideline adherence:} Correctly applies recommendations for nitrofurantoin use in pregnancy (safe in second trimester, avoided in first/third).
\end{itemize}

\subsubsection{Reasoning Process}
The reasoning follows a hypothetico-deductive model common in clinical medicine:
\begin{itemize}
    \item \textbf{Information gathering:} Patient demographics, symptoms, vital signs, and exam findings.
    \item \textbf{Problem representation:} “Pregnant woman with dysuria, no systemic signs, likely cystitis.”
    \item \textbf{Differential diagnosis:} Prioritizes cystitis over pyelonephritis.
    \item \textbf{Treatment selection:}
    \begin{itemize}
        \item \textbf{Elimination:} Doxycycline (contraindicated).
        \item \textbf{Comparison of remaining options:} Ampicillin (resistance), ceftriaxone (overly broad), nitrofurantoin (guideline-supported).
        \item \textbf{Final decision:} Nitrofurantoin, justified by safety in the second trimester and efficacy for uncomplicated cystitis.
    \end{itemize}
\end{itemize}

We believe that the structured reasoning approach with high accuracy shows the usefulness of DeepSeek R1 in the healthcare sector. The sound reasoning combines with an open source model gives a clear path forward for integrating this in the healthcare domain.

\section{Discussion}
This study provides a detailed analysis of the medical reasoning capabilities of DeepSeek R1, revealing both its strengths and limitations in handling complex clinical scenarios. While the model demonstrates high overall diagnostic accuracy (93\%), our in-depth error analysis highlights specific areas where its reasoning leads to errors in clinical assessment. These findings have several important implications for the development and deployment of LLMs in healthcare.

\subsection{A note on anthropomorphization of LLMs}
In this work we evaluated the reasoning of LLMs and highlighted cognitive errors in its reasoning. There is a speculative nature to this since we assign human error mechanism to an LLM system. We want to be clear that the bias we found in reasoning is dependent on the analysis of the reasoning text and we provide all model reasoning outputs as supplementary material. The language we use to describe the reasoning and errors is made to help human understanding and we hope that this does not lead to anthropomorphization of these systems. We believe that LLMs should be viewed as tools but language regarding human cognition can help increase our understanding of their functioning.

\subsection{Opening the black box}
Deep learning models including LLMs have been accused of being black box algorithms where the inner workings of the models are shielded from view\cite{wang2024augmentingblackboxllmsmedical}. This has limited their use in high risk areas such as healthcare where understanding of model outputs is essential for safe implementation. Open reasoning models such as R1 shows a path forwards by being transparent regarding reasoning which has the potential of making the model safer to use in a a high risk setting.

\subsection{Errors in medical reasoning}
Errors that took place were overall a result of thinking errors where the model focused too much attention on details of a problem and lacked necessary understanding of medical protocols. These errors can be viewed similar to mistakes made by a human with medical knowledge and ability to reason about that knowledge making a mistake. i.e. a doctor misdiagnosing a patient rather than a human without medical knowledge guessing the answer on a medical test. This is an important distinction because the difference between the two is years of clinical schooling and medical reasoning ability. As such we view these errors as promising and believe that training techniques and new reasoning models will enhance this already fairly adequate medical reasoning ability. Our findings that the length of reasoning was strongly linked to correctness is interesting and can be helpful for improving the usefulness of these models in a clinical setting. By simply using the length of reasoning as a reverse certainty score, we can help a clinician make sense of the models reasoning and even automate double checking, by rerunning long reasoning attempts with an added prompt that the reasoning is likely incorrect.

\subsection{Quality of Medical Reasoning}
Overall we found that the model made few mistakes in its reasoning and the reasoning was medical in nature. The model could reason regarding medical scenarios and  overall the reasoning of the model was excellent. This is promising because it shows that medical reasoning is possible through LLMs and that the reasoning is already functional and can be helpful in the healthcare sector if integrated in a safe way. 

\subsection{The future of LLMs in healthcare}
As within other areas of healthcare, expert clinicians time become a bottleneck when evaluating LLMs. As models improve and show signs of medical reasoning it seems worthwhile to use LLMs to improve LLMs in healthcare. This seemingly paradoxical way of working is actually in line with how large AI labs work to improve LLMs\cite{anthropic2023claude}. A capable LLM model can be used to refine and improve data that can be used to train another LLM and over time data quality improves as well as model performance. For larger medical datasets where human evaluation is simply unfeasible when thousand or millions of questions are evaluated this technique becomes necessary. Having a gold standard of human evaluation with lesser standards for evaluation using LLMs seems to be a possible way forward. As in other areas where LLMs are highly performant such as code generation, we should start to accustom ourself to a world where clinicians supervise AI systems that reason independently. In the future the job of the clinician might be to supervise an AI system that independently gives suggestions for diagnosis and treatment.

\subsection{Improving human medical reasoning}
Errors in medical reasoning by humans leads to thousands of deaths and injuries each year \cite{Makaryi2139}. As such improving clinicians ability to reason might be one of the most important tasks for improving healthcare outcomes. The medical reasoning already available in the R1 model can take years for a clinician to acquire through medical training and mentorship and thus using models such as R1 to improve clinicians reasoning skills is one potential use of this technology. This is also in line with a human in the loop approach which improves safety while being aligned with regulatory bodies views on AI in healthcare \cite{eu20241689}.

\subsection{Improving clinical reasoning}
The model was evaluated with a simple prompt and could likely improve through several methods.

\begin{enumerate}
    \item Retrieval augmented generation (RAG) for improved clinical reasoning. By using a RAG system the performance of the system would likely improve by access to clinical guidelines and other medical texts.
    \item Specialization in prompting and documents. In a clinical context, medical professionals usually reason about a smaller subset of clinical knowledge. By dividing the problem of medical reasoning by medical specialty; prompts and knowledge could be used to solve these subproblem more appropriately.
    \item Fine tuning on medical reasoning. Improvements to medical reasoning would likely result from fine-tuning on medical reasoning data. Recent advancements in reinforcement learning training for text \cite{deepseekai2025deepseekr1incentivizingreasoningcapability} could be useful in this regard.
\end{enumerate}

\subsection{Use in a clinical setting}
Although the model had errors, overall the reasoning was sound from a medical perspective, as such we believe that these models can be useful in the medical domain and we think it is time for healthcare practitioner to start experimenting with these technologies. As long as healthcare workers are aware of limitations, we believe that use of these systems could help improve patient outcomes. For many clinicians especially in specialized care settings the work can be lonely and there might not be colleagues with similar experience to discuss medical diagnostics. Even though healthcare decisions should always be the responsibility of a human, we believe that reasoning models such as R1 can help clinicians in their diagnostic assessments.

As clinicians we need to be creative in finding safe ways to use this technology in a clinical settings. Both for clinician facing and patient facing interfaces there are likely useful ways to use this technology in a way that is helpful for improving health outcomes.

\subsection{Limitations}
This study has several limitations. First, the evaluation is based on a limited, albeit diverse, set of clinical cases from a single dataset. While MedQA provides a valuable benchmark, it may not fully capture the complexity of real-world clinical practice. Second, our analysis focuses on one specific LLM, DeepSeek R1. While this model represents a state-of-the-art approach to reasoning-enhanced LLMs, the findings may not be generalizable to all LLMs, especially those with different architectures or training methodologies. Third, the expert validation is still subject to the inherent limitations of human judgment and potential biases. Another limitation is that we only had a single medical expert evaluate the medical reasoning of the model.

\subsection{Future Research Directions}
Future research should focus on developing more robust evaluation frameworks that encompass a broader range of clinical scenarios and incorporate dynamic, real-time interactions. Investigating the effectiveness of different prompting strategies, retrieval augmented generation and fine-tuning methods in improving reasoning performance is also crucial. Furthermore, exploring hybrid AI-clinician collaborative models, where LLMs serve as decision support tools rather than autonomous diagnostic agents, could leverage the strengths of both human and artificial intelligence.

\subsection{Conclusion}
This study shows that DeepSeek R1 is capable of a form of medical reasoning as evaluated by analysis by human evaluation on a subset (n=100) of the MedQA benchmark. The model had an accuracy of 93\% and both correct and incorrect cases showed signs of medical reasoning. Using open reasoning models in healthcare improves explainability over non-reasoning models and we encourage continued investigation of how these models can be used to improve the future of healthcare.

\bibliographystyle{plainnat}
\bibliography{references}

\end{document}